\documentclass{article}
\usepackage{graphicx}
\usepackage{amsmath, amssymb}
\usepackage{authblk}
\usepackage{booktabs}
\usepackage{hyperref}
\usepackage[margin=1in]{geometry}
\usepackage{graphicx}    
\usepackage{caption}  
\usepackage{subfigure}  
\usepackage{float}

\title{PM25Vision: A Large-Scale Benchmark Dataset for Visual Estimation of Air Quality}

\author[1]{Han Yang}
\affil[1]{Department of Computer Science, New York University}

\begin{document}

\maketitle

\begin{abstract}
We introduce PM25Vision (PM25V), the largest and most comprehensive dataset to date for estimating air quality—specifically PM2.5 concentrations—from street-level images. The dataset contains over 11,114 images matched with timestamped and geolocated PM2.5 readings across 3,261 AQI monitoring stations and 11 years, significantly exceeding the scale of previous benchmarks. The spatial accuracy of this dataset has reached 5 kilometers, far exceeding the city-level accuracy of many datasets. We describe the data collection, synchronization, and cleaning pipelines, and provide baseline model performances using CNN and transformer architectures. Our dataset is publicly available.
\end{abstract}

\section{Introduction}

Air pollution is a critical environmental issue that significantly impacts human health and ecological stability. Traditional air quality monitoring systems, particularly for pollutants like PM2.5 and PM10, rely heavily on fixed monitoring stations. However, these stations are often sparsely distributed, providing limited spatial resolution and failing to capture fine-grained variations in urban environments. This is particularly true in developing countries, where comprehensive Air Quality Index (AQI) monitoring systems are often absent or underdeveloped.

In recent years, there has been an increase in publicly available web and camera images, such as street photos or CCTV pictures, which has made AQI estimation based on ground-based imagery (as opposed to remotely sensed imagery) possible. These images often visibly reflect environmental conditions, including haze density, visibility, and sky clarity—factors closely associated with air pollution levels.

In this context, a variety of studies have emerged on estimating AQI based on smartphones, fixed cameras, and more. These studies used everything from rule-based to complex attention networks for modeling, yielding many useful results. These achievements can serve as a supplementary means for PM2.5 estimation and play a significant role in less developed areas. However, current research on computer vision (CV) prediction of AQI is still scattered and independent. Most studies use self-developed datasets, which are usually limited to the cities or countries where the authors are located, and have limited spatial accuracy, data volume and image quality. 

Therefore, this study created PM25Vision --- a global high-quality large-scale dataset, which was achieved by matching data from the current largest air quality project World Air Quality Index Project (WAQI)\cite{waqi} and the largest crowdsourced street view photo platform Mapillary\cite{mapillary}. This study downloaded historical PM2.5 data from approximately 13,890 WAQI monitoring stations worldwide, as well as 515,961 photos on mapillary that could be paired with PM2.5 data. Through a complete data cleaning pipeline, a final high-quality, balanced 11,114 images and their PM2.5 information were obtained. As the final dataset. 

This study also used some common CV models for benchmarking on the dataset. PM25V dataset is currently the state-of-the-art AQI image prediction dataset and is expected to become the benchmark dataset in this field, enhancing the prediction ability and universality of existing AQI-focused CV models.

\section{Related Work}
Research on the field of image-based AQI estimation is relatively sparse and independent. This section will introduce those studies with clear and reproducible methodologies.

Since there is a greater demand for this technology in developing countries, research has focused on countries such as China, India, and Bangladesh. A study from Hangzhou, China utilized low-altitude data collected by drones to train machine learning models such as RF and SVR\cite{Hangzhou}. They constructed two effective, physics-based features: Dark Channel Prior (DCP) and Standard Deviation of Grayscale. The intuitive meaning of DCP is that light scattering caused by PM2.5 makes dark regions appear brighter, while the Std of Grayscale means that PM2.5 makes the image blurrier, reducing the standard deviation. A study from Lanzhou, China, used five fixed cameras and made predictions via ResNet combined with Spatial and Context Attention block, also known as AQC-Net, which gave slightly better results than the default ResNet18\cite{AQCnet}.

A study from India predicted AQI from pictures obtained from in-car cameras, also found that using AQC-Net gave the best results\cite{IndiaTraffic}. Another study from India collected about 12,240 street view images from India and Nepal and used VGG16 as a model to make predictions\cite{inne1, inne2}. 
A study from Bangladesh used 1818 street view images of Dhaka city uploaded by citizens to predict PM2.5 concentration by a CNN network, Deep Convolutional Neural Network (DCNN), which is designed to be more lightweight and got better results than networks such as ViT, ResNet50\cite{Bang}.

\section{Dataset Construction}
\subsection{Data Sources}
We construct our dataset by integrating two large-scale, publicly accessible platforms: WAQI and Mapillary. WAQI provides historical daily AQI data from thousands of air quality monitoring stations worldwide. Mapillary is a global street-level imagery platform containing millions of geo-tagged and time-stamped images contributed by users and organizations. 

\subsection{Data Pairing}
We first download all PM2.5 AQI history csv files in a folder and get all station's longitude and latitude data via WAQI's API. There are 13,890 files corresponding to 13,890 stations. We filter out csv files with abnormal or insufficient data through indicators such as variance. After this step, we still have files from 11,844 monitoring stations. After that, we traversed each site, using the Mapillary API to search for images within 5 kilometers of the station's location, and checked whether there was an AQI value corresponding to the image date in the historical data. If there was, the image and the corresponding PM2.5 information were stored in the dataset folder. There are large number of continuous pictures taken on a section of road on Mapillary, to ensure diversity of the photos, we only paired up to 100 pictures for each monitoring station. Completing this process let us to obtain 515,961 images.

\subsection{Data Cleaning Pipeline}
\begin{figure}[H]
    \centering
    \subfigure[Good]{
        \includegraphics[width=0.45\linewidth]{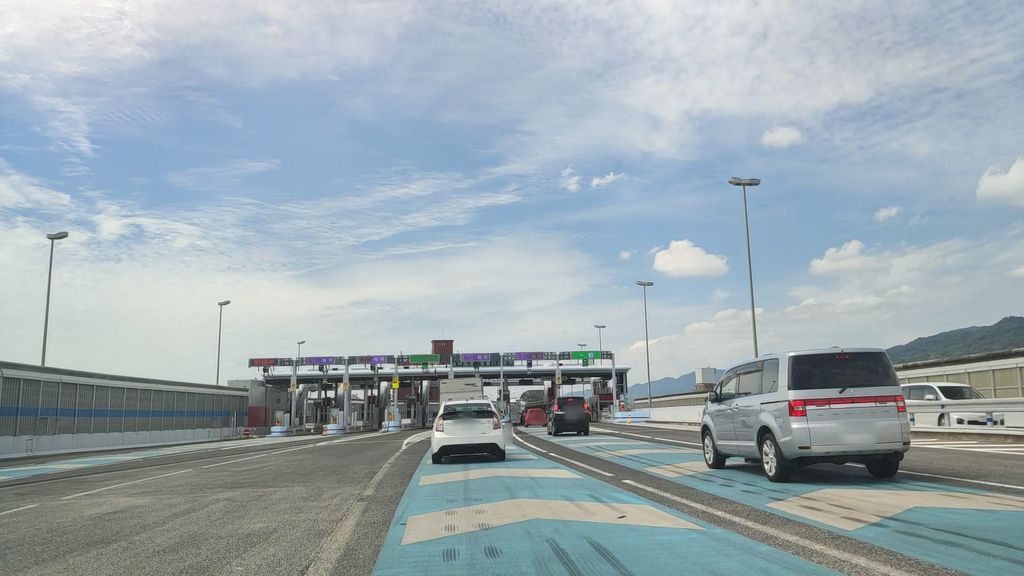}
    }
    \hspace{0.03\linewidth} 
    \subfigure[Bad]{
        \includegraphics[width=0.45\linewidth]{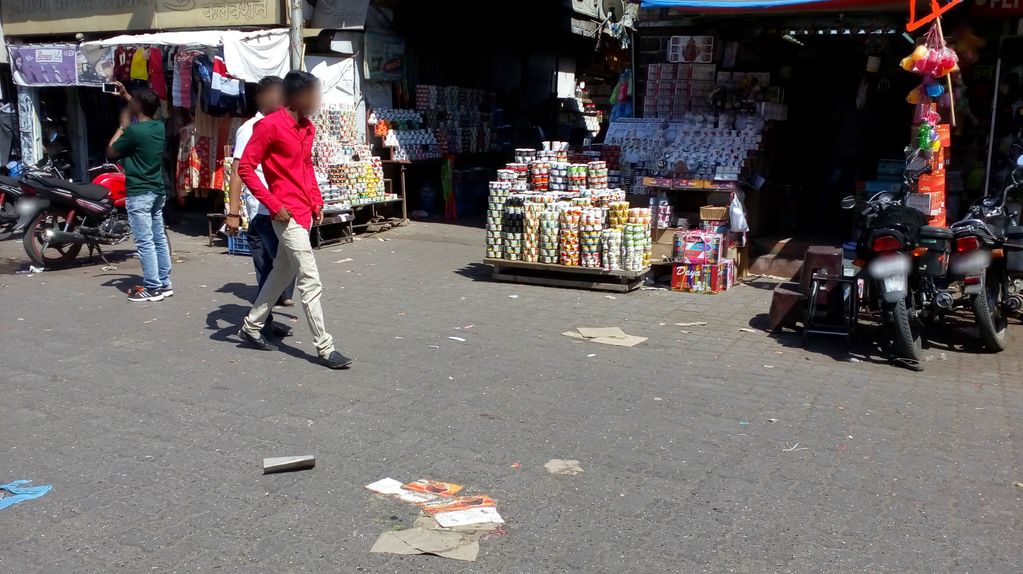}
    }
    \caption{"Good" and "bad" images for PM2.5 predictions.}
    \label{fig:gab}
\end{figure}

We removed the paired data where PM2.5 is a NaN value. 
Some images in Mapillary are not suitable for PM2.5 prediction, such as those without sky, overexposed, and at night, as shwon in \autoref{fig:gab}. Existing CV models have difficulty determining which images are "good" for PM2.5 predcition and which are "bad". To address this, we manually classified 500 photos (good, bad, and uncertain), divided the training and test sets in an 8:2 ratio, and then let a ResNet18\cite{he2015} learn the manual classification (only good and bad). It quickly converged and achieved an accuracy rate of 95\% on the test set. We used it to filter out the bad images, which accounted for approximately one-third of the dataset. 

\begin{figure}[H]
    \centering
    \includegraphics[width=0.6\linewidth]{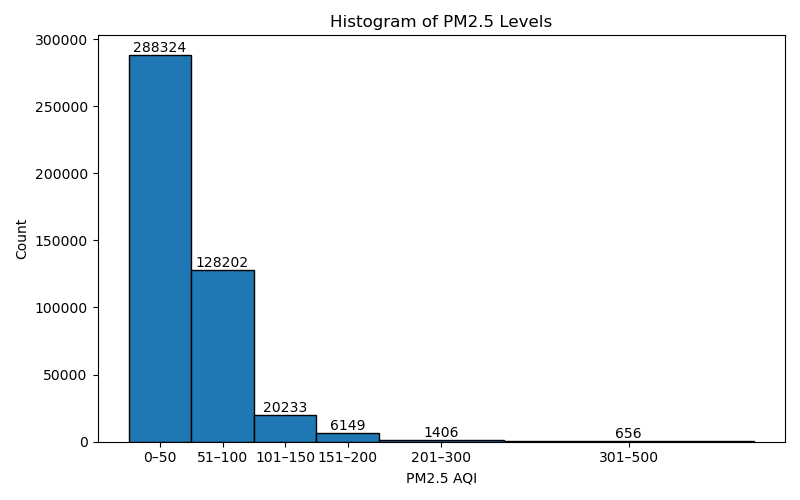}
    \caption{Histogram that shows the class imbalance problem.}
    \label{fig:hist}
\end{figure}

\begin{figure}[H]
    \centering
    \includegraphics[width=0.6\linewidth]{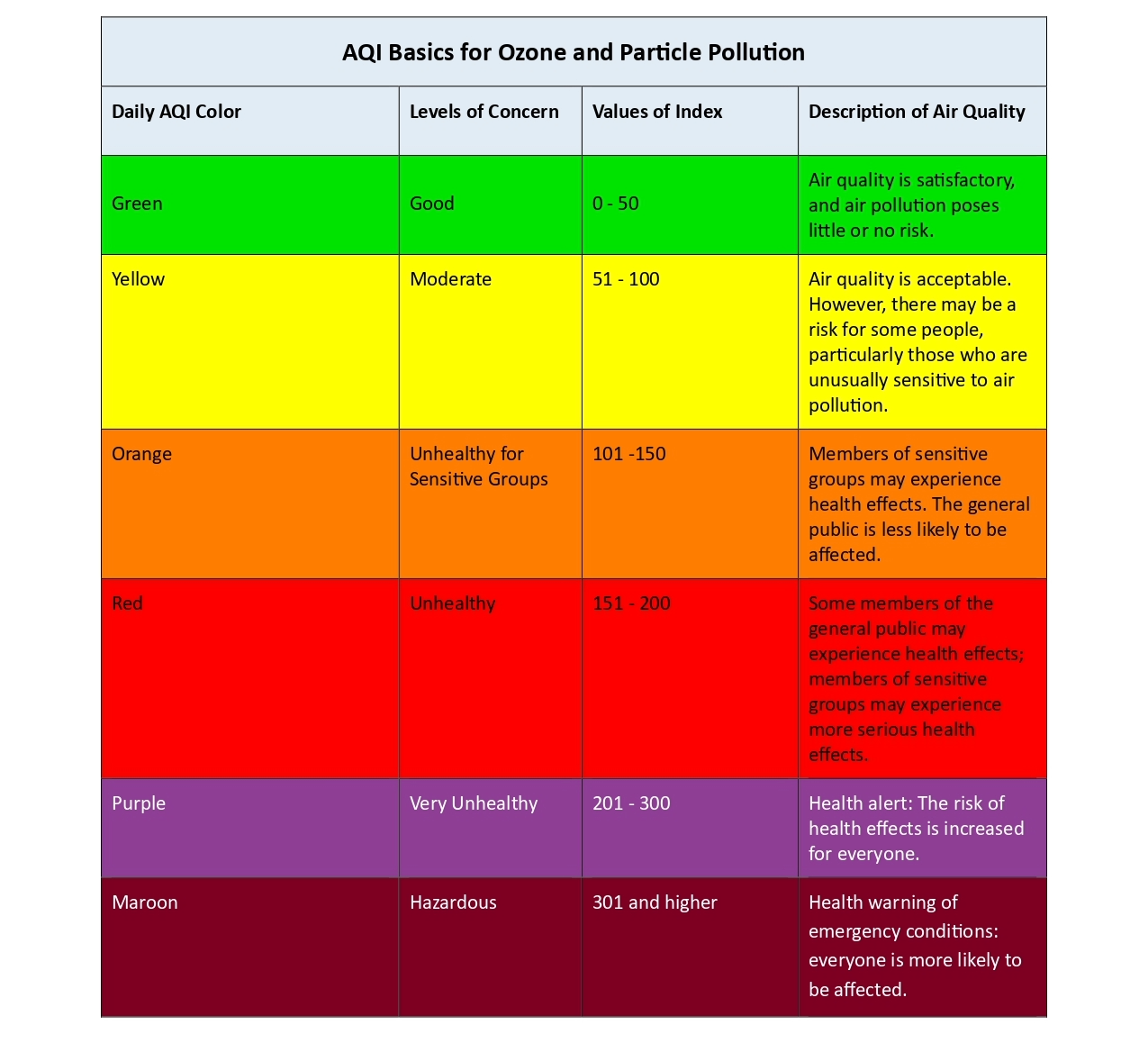}
    \caption{Introduction of AQI levels.}
    \label{fig:aqi_levels}
\end{figure}

Then, we divided the dataset into the training and test sets in an 8:2 ratio. Due to the fact that the photos in Mapillary are concentrated in developed areas or deliberately taken during fine weather for other CV tasks, this leads to a serious class imbalance problem in the data, as shown in \autoref{fig:hist}. Note that our AQI classification is consistent with US Environmental Protection Agency, as shown in \autoref{fig:aqi_levels}. Therefore, we performed sample balancing on the train and test sets respectively. The balancing rule is that the number of samples in the class with the largest number of samples cannot exceed five times that of the minimum class. This process removed the vast majority of the photos, and the final data size is 11,114 images.

\subsection{Statistics}
\begin{figure}[H]
    \centering
    
    \includegraphics[width=0.30\linewidth]{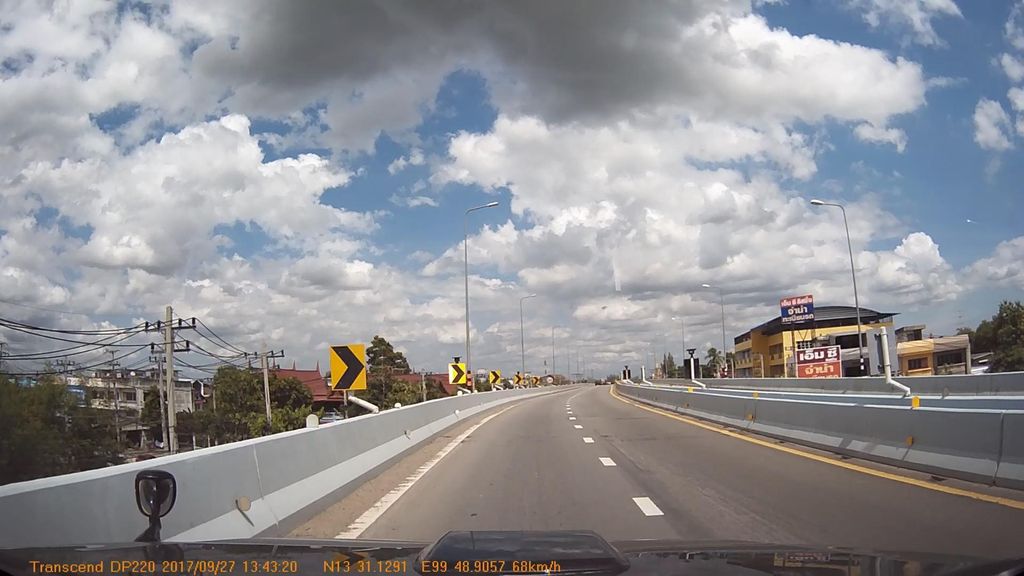}
    \hspace{0.02\linewidth}
    \includegraphics[width=0.30\linewidth]{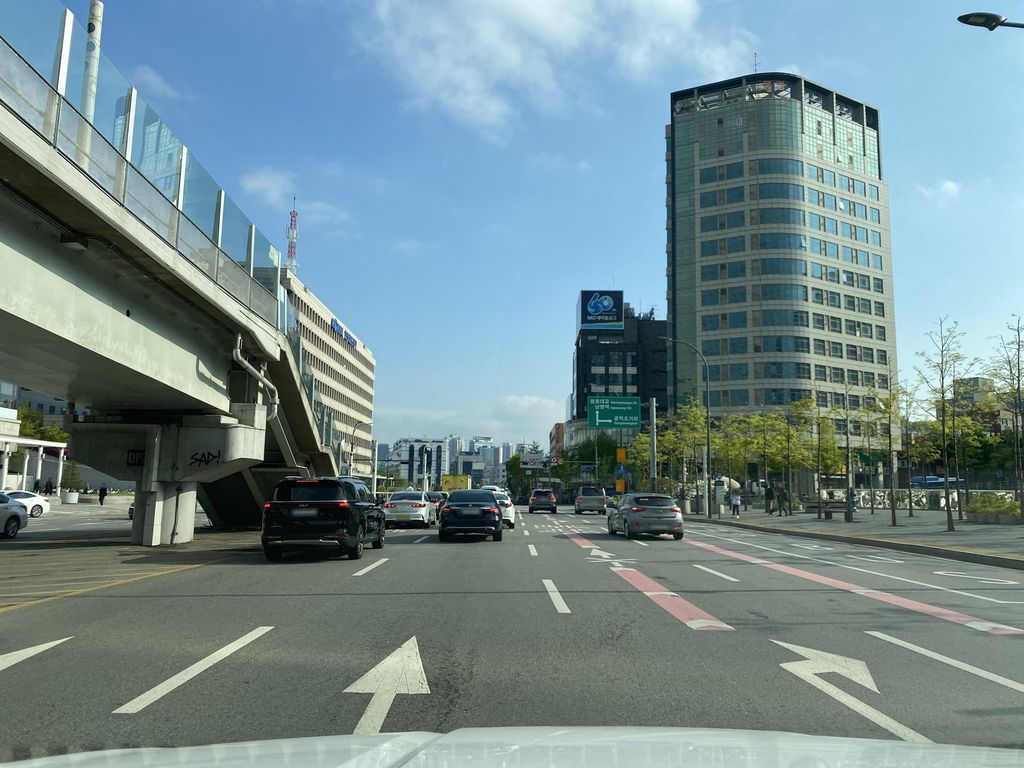}
    \hspace{0.02\linewidth}
    \includegraphics[width=0.30\linewidth]{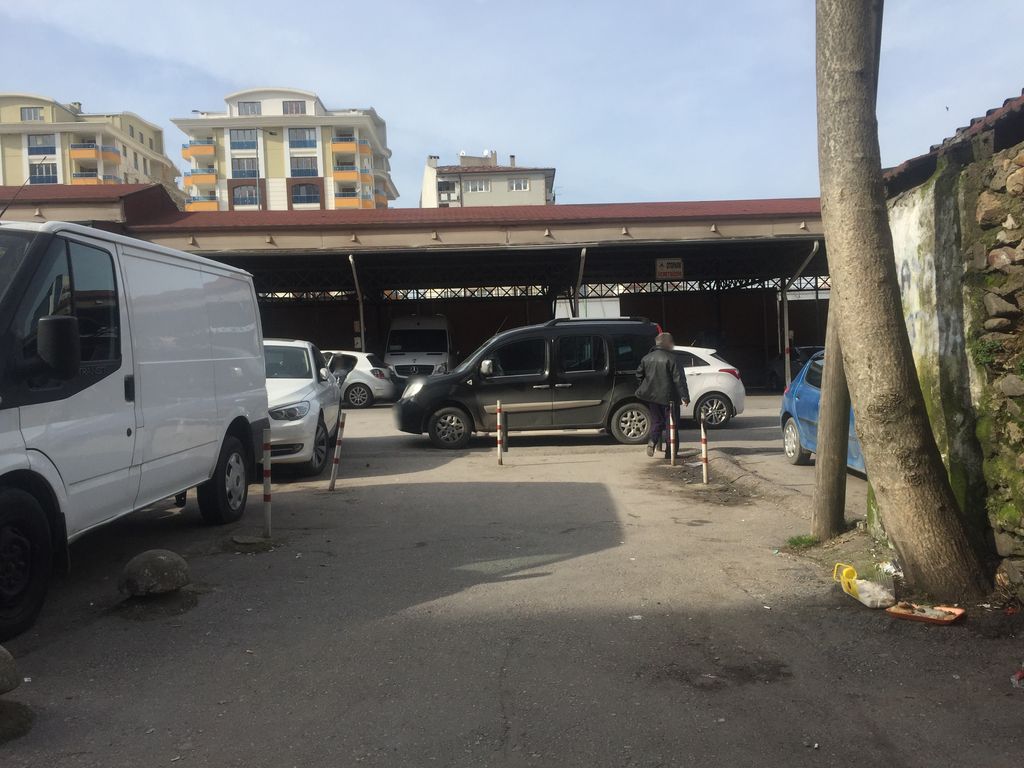}

    \vspace{0.02\linewidth}

    \includegraphics[width=0.30\linewidth]{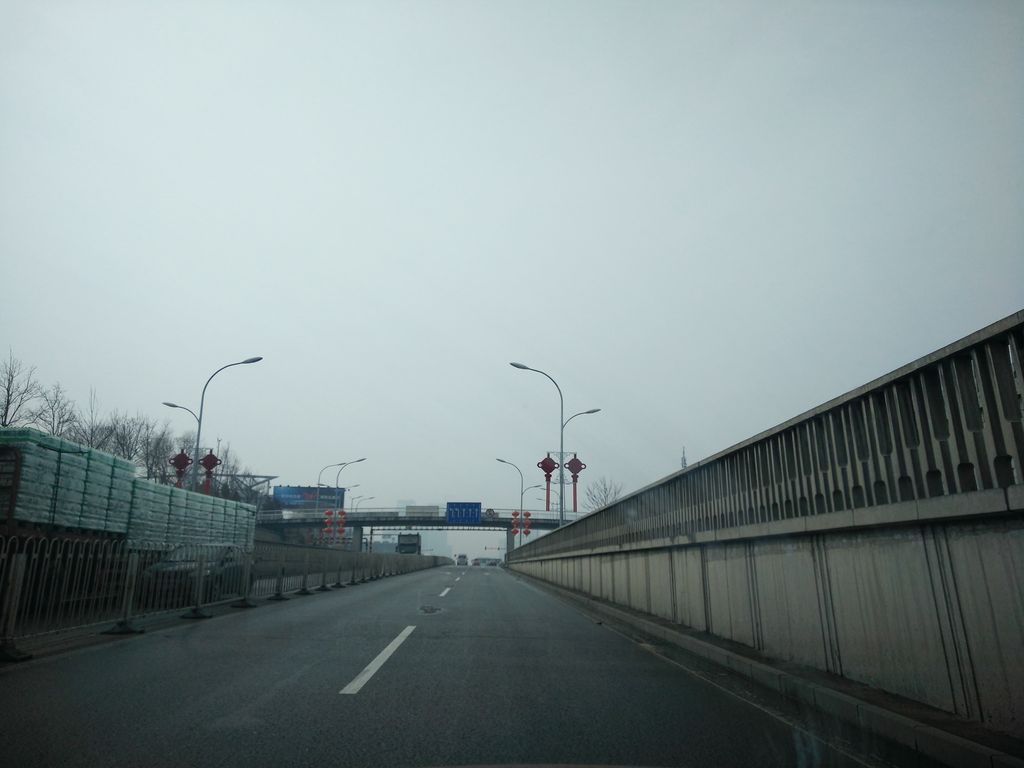}
    \hspace{0.02\linewidth}
    \includegraphics[width=0.30\linewidth]{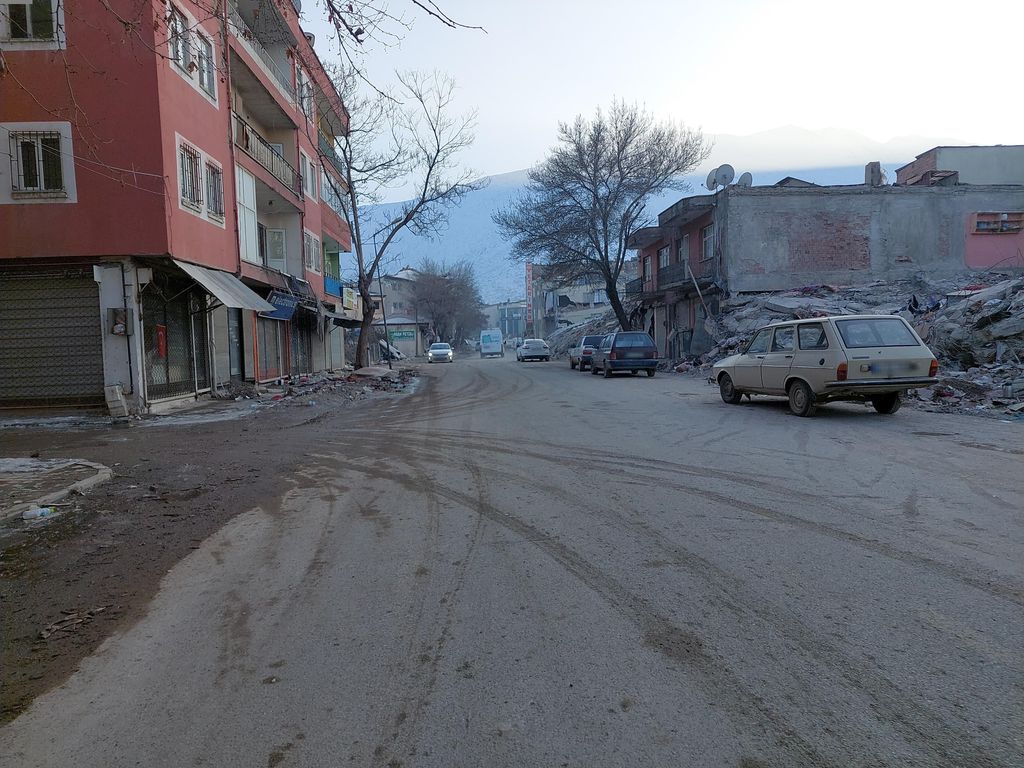}
    \hspace{0.02\linewidth}
    \includegraphics[width=0.30\linewidth]{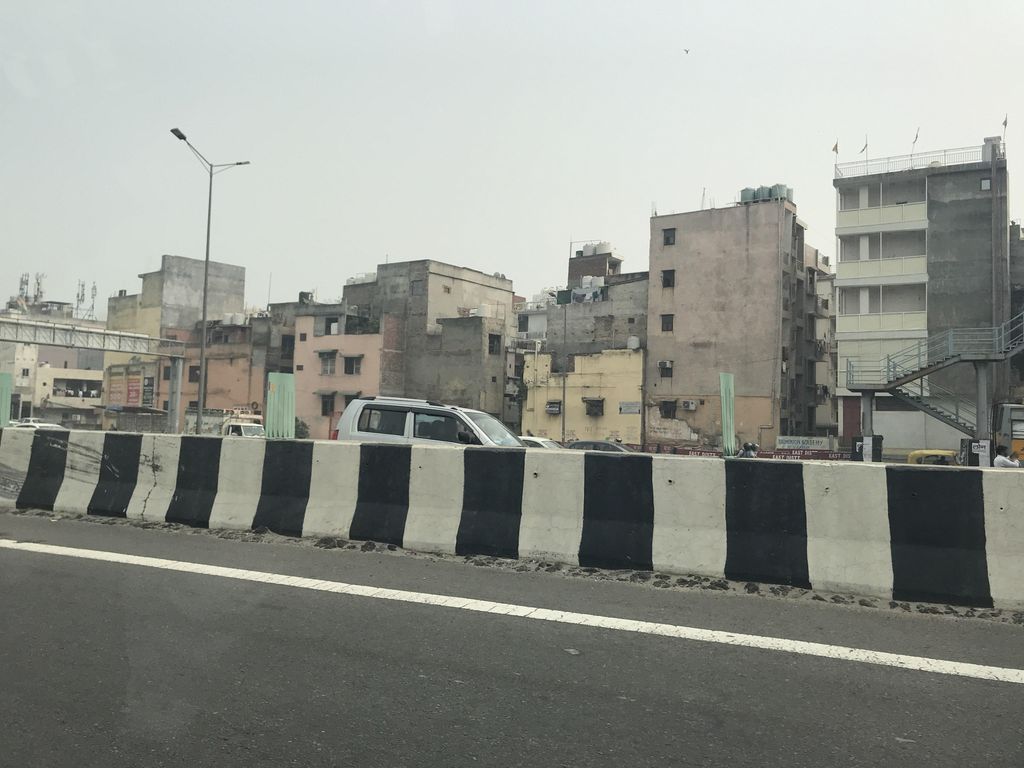}

    \caption{Example images from the PM25Vision dataset representing classes 0 through 5.}
    \label{fig:pm25v_examples}
\end{figure}

The final PM25Vision dataset contains 11,114 images around 3,261 different stations, spanning the years 2014 to 2025. One can see the diversity from \autoref{fig:pm25v_examples}.

\section{Baseline Models \& Results}
In this study, we trained three representative CV models—EfficientNet-B0\cite{tan2020}, ResNet50, and ViT-B/16\cite{dosovitskiy2021}—on our dataset. Each model was trained for both regression and classification tasks: the regression task aimed to predict the exact AQI value, while the classification task focused on predicting the corresponding AQI level. The model performance on regression task is shown in \autoref{tab:regression_perf}. Note that we also reported classification metrics by discretizing the regression outputs into AQI levels. The model performance on classification task is shown in \autoref{tab:classification_perf}.

\begin{figure}[H]
    \centering
    \subfigure[Loss curves]{
        \includegraphics[width=0.48\linewidth]{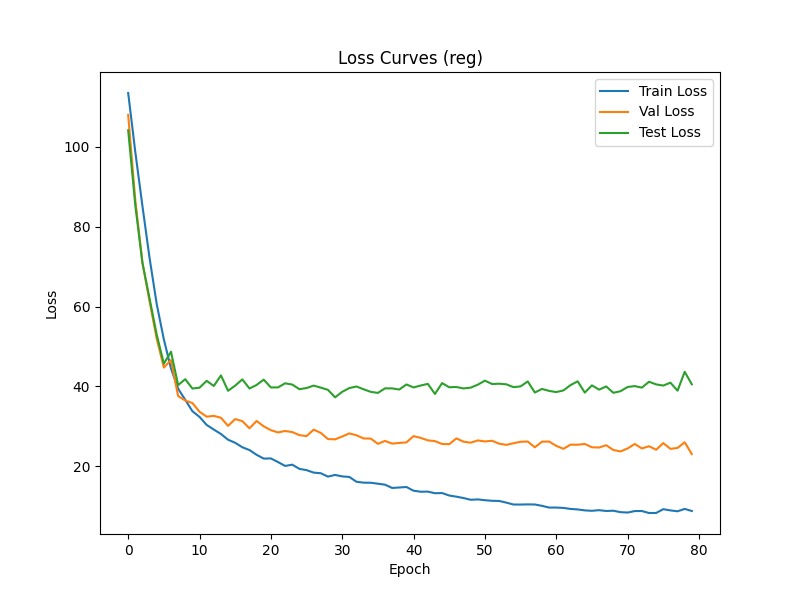}
    }
    \hspace{0.01\linewidth} 
    \subfigure[Confusion matrix]{
        \includegraphics[width=0.42\linewidth]{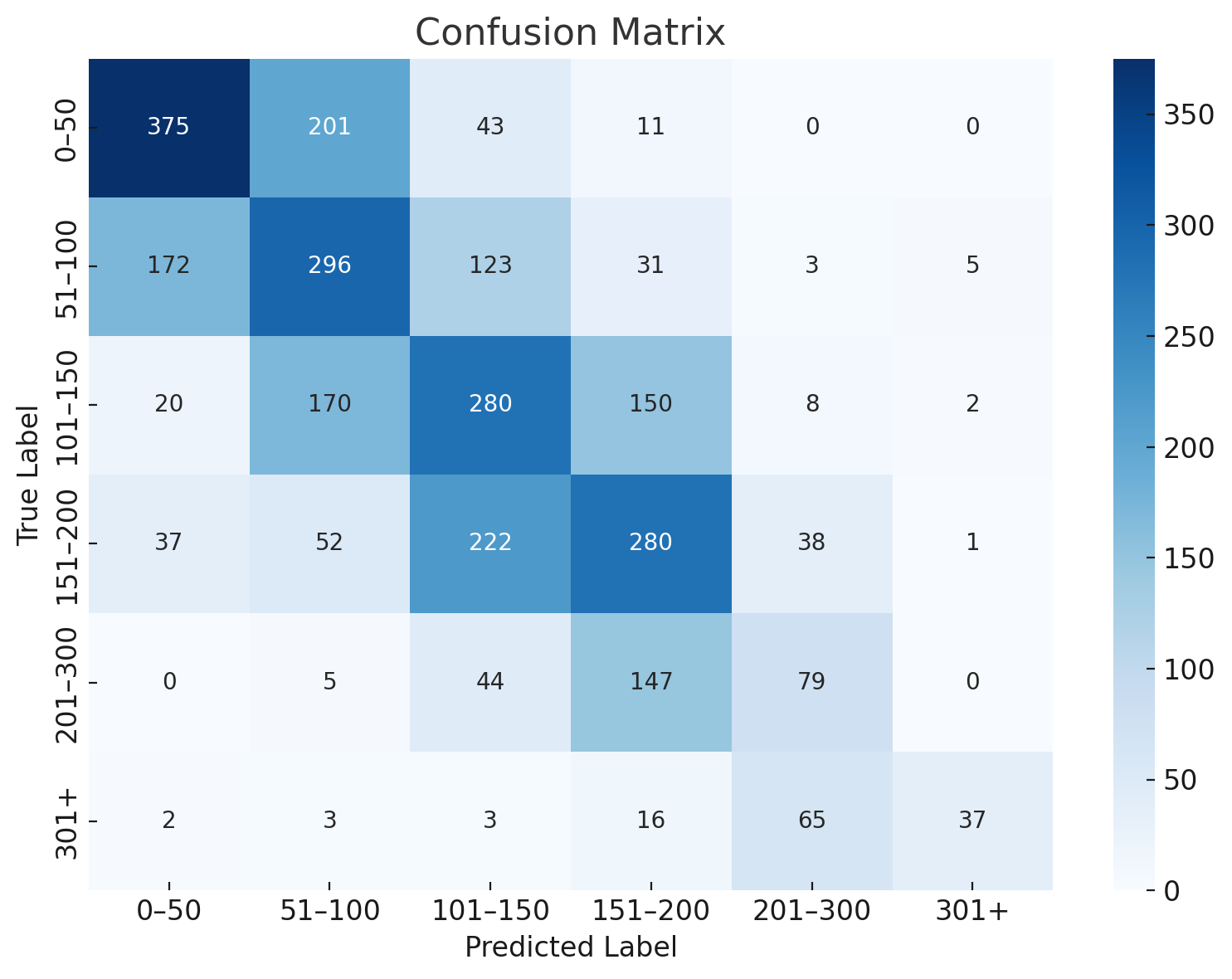}
    }
    \caption{Training loss curves and confusion matrix of EfficientNet-B0 on regression task.}
    \label{fig:train_detail}
\end{figure}

EfficientNet-B0 achieves the best performance on the regression task, with the highest R² (0.550), lowest RMSE (54.6) and highest F1 score (0.445), indicating strong predictive accuracy for continuous AQI values.
While ResNet50 slightly outperforms it on the classification task, EfficientNet-B0 still shows competitive classification performance when evaluated on discrete outputs. Overall, EfficientNet-B0 demonstrates the best overall performance among the three models, see \autoref{fig:train_detail} for more information. We can also see that ViT-B/16 underperforms, possibly due to its transformer-based design, which may not suitable for air quality prediction.

\begin{table}[H]
\centering
\caption{Performance of regression models evaluated with both regression and classification metrics.}
\label{tab:regression_perf}
\begin{tabular}{l|ccc|ccc}
\toprule
\textbf{Model} & R$^2$ ↑ & MAE ↓ & RMSE ↓ & Acc ↑ & F1 ↑ & Precision ↑ \\
\midrule
EfficientNet-B0 & 0.550 & 36.6 & 54.6 & 0.461 & 0.445 & 0.515 \\
ResNet50        & 0.502 & 38.6 & 57.5 & 0.436 & 0.348 & 0.446  \\
ViT-B/16        & 0.225 & 50.3 & 71.7 & 0.348 & 0.296 & 0.403  \\
\bottomrule
\end{tabular}
\end{table}

\begin{table}[H]
\centering
\caption{Performance of classification models on AQI level prediction.}
\label{tab:classification_perf}
\begin{tabular}{lccccc}
\toprule
\textbf{Model} & Accuracy ↑ & F1 ↑ & Precision ↑ & Recall ↑ & Balanced Acc ↑ \\
\midrule
EfficientNet-B0 & 0.399 & 0.336 & 0.420 & 0.332 & 0.332 \\
ResNet50        & 0.440 & 0.380 & 0.476 & 0.372 & 0.372 \\
ViT-B/16        & 0.402 & 0.366 & 0.412 & 0.361 & 0.361 \\
\bottomrule
\end{tabular}
\end{table}

\section{Limitations \& Future Work}
Severe class imbalance limits the sample size of the PM25Vision dataset, which can be improved in the future through a more refined mapillary image download method, as there are a large number of images with poor air quality on mapillary. 

In addition, the temporal resolution of historical data on WAQI is limited to the day level. Due to in-day fluctuations in air quality, this may lead to some inaccurate annotations. Compared with the spatial resolution of approximately 5 kilometers, temporal accuracy is the primary factor affecting the quality of our dataset.

Finally, this dataset only uses PM2.5 as label, but the WAQI contains AQI data for other air pollutants such as PM10 and O3, which is worthy of further study.

\section{Conclusion}
We presented PM25Vision, a large-scale benchmark dataset for visual estimation of air quality, constructed by aligning Mapillary street-level images with WAQI PM2.5 records. Through careful filtering and balancing, we obtained a clean and diverse dataset of 11,114 images spanning more than a decade and over 3,000 monitoring stations worldwide. 

Baseline experiments with representative convolutional and transformer models show that convolutional networks, particularly EfficientNet-B0, achieve the best regression performance. We believe PM25Vision provides a solid foundation for future research on vision-based environmental research and can inspire more advanced models and applications in this emerging field.

\section{Acknowledgements}
We would like to thank Mapillary for providing crowdsourced street-level images, and the World Air Quality Index Project (WAQI) for open air quality data. We are additionally grateful to the U.S. Embassy in Beijing for publishing PM2.5 data since 2011, which raised public awareness of air pollution in China and motivated subsequent research efforts, including this study.

\section{Availability}
The full dataset is available at \url{https://huggingface.co/datasets/DeadCardassian/PM25Vision} and \url{https://www.kaggle.com/datasets/deadcardassian/pm25vision}. We also deployed an online PM2.5 image prediction applicaiton based on our dataset and the best-performing model, available at \url{www.pm25vision.com}. The codes for original data downloading, processing and model training are available upon request.

\bibliographystyle{unsrt}   
\bibliography{references}

\end{document}